\documentclass[journal]{IEEEtran}
\usepackage{algorithm,algorithmic}
\usepackage{graphicx}
\usepackage{amsmath}
\usepackage{lipsum}
\usepackage{booktabs}
\usepackage{algorithmic}
\usepackage{multirow}
\usepackage[acronym]{glossaries}
\usepackage{cite}
\usepackage{flushend}
\usepackage[bottom]{footmisc}
\usepackage{perpage} %the perpage package

\begin{document}
\title{Intelligent \textit{Known} and \textit{Novel} Aircraft Recognition - A Shift from Classification to Similarity Learning for Combat Identification}
\author{\IEEEauthorblockN{ Ahmad Saeed\textsuperscript{a}, Haasha Bin Atif\textsuperscript{a}, Usman Habib\textsuperscript{a}, Mohsin Bilal\textsuperscript{a,b} }
\\
\text{\textsuperscript{a}Department of Artificial Intelligence \& Data Science }
\\
\text{National University of Computer and Emerging Sciences, Islamabad, Pakistan}
\\
\text{\textsuperscript{b}College of Computer Engineering and Sciences }
\\
\text{Prince Sattam Bin Abdulaziz University, Al-Kharj, Saudi Arabia}
\\
\IEEEauthorblockA{\textit{\{I212281, haasha.atif, usman.habib, mohsin.bilal\}@nu.edu.pk}}

}

\maketitle

\begin{abstract}

Precise aircraft recognition in low-resolution remote sensing imagery is a challenging yet crucial task in aviation, especially combat identification. This research addresses this problem with a novel, scalable, and AI-driven solution. The primary hurdle in combat identification in remote sensing imagery is the accurate recognition of \textit{Novel}/\textit{Unknown} types of aircraft in addition to \textit{Known} types. Traditional methods, human expert-driven combat identification and image classification, fall short in identifying \textit{Novel} classes. Our methodology employs similarity learning to discern features of a broad spectrum of military and civilian aircraft. It discerns both \textit{Known} and \textit{Novel} aircraft types, leveraging metric learning for the identification and supervised few-shot learning for aircraft type classification.To counter the challenge of limited low-resolution remote sensing data, we propose an end-to-end framework that adapts to the diverse and versatile process of military aircraft recognition by training a generalized embedder in fully supervised manner.Comparative analysis with earlier aircraft image classification methods shows that our approach is effective for aircraft image classification (F1-score Aircraft Type of $0.861$) and pioneering for quantifying the identification of \textit{Novel} types (F1-score Bipartitioning of $0.936$). The proposed methodology effectively addresses inherent challenges in remote sensing data, thereby setting new standards in dataset quality. The research opens new avenues for domain experts and demonstrates unique capabilities in distinguishing various aircraft types, contributing to a more robust, domain-adapted potential for real-time aircraft recognition.

\end{abstract}


\section{Introduction}
\begin{figure}[ht]
  \centering
  \includegraphics[width=0.9\linewidth]{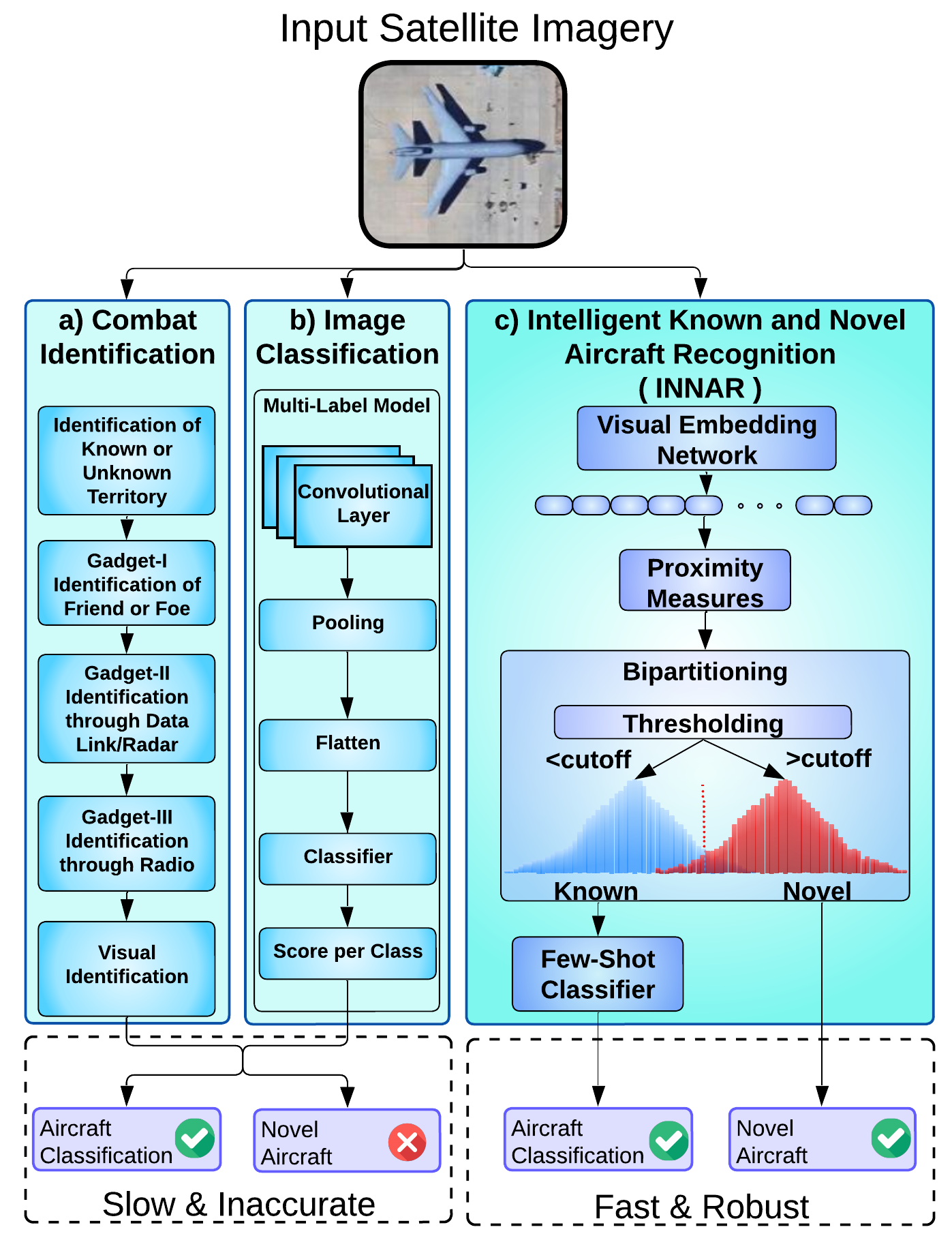}
  \caption{Flow Diagram from left to Right showing  a)Traditional CID , b) Image classification \& c)  Intelligent Known and Novel Aircraft Recognition (INNAR) to find \textit{Known} and \textit{Novel} class }
  \label{fig:introMethodology}
\end{figure}
Aircraft recognition in remote sensing imagery presents a far-reaching challenge due to its wide-ranging applications, encompassing border security surveillance, emergency response, disaster management, air traffic control, environmental monitoring, maritime surveillance, counter-terrorism efforts and critical asset security. The principal challenge of combat identification (CID)~\cite{b51:andrews2012human,b46:tan2023research} stems from the need to identify and label potential targets as friend, foe, or neutral aircraft, a task of considerable practical importance. The primary hurdle in CID is the accurate recognition of \textit{Novel}/Unknown classes of aircraft in addition to \textit{Known} aircraft types (or classes). Recognition hinges on distinguishing \textit{Known} classes from \textit{Novel} classes (unfamiliar/disjoint types). The former represents established types encountered during training, while the latter poses challenges due to the model's lack of explicit training. Traditional methods, such as human expert-driven CID (Figure \ref{fig:introMethodology}-a, and image classification-based aircraft recognition (Figure \ref{fig:introMethodology}-b, fall short in identifying \textit{Novel} classes. This paper presents a pioneering methodology termed '\textbf{I}ntelligent k\textbf{N}own and \textbf{N}ovel \textbf{A}ircraft \textbf{R}ecognition (INNAR)' illustrated in Figure \ref{fig:introMethodology}-c. This innovative approach signifies a paradigm shift from traditional image classification to similarity learning, with the objective of automating CID. It facilitates the recognition of both \textit{Known} and \textit{Novel} aircraft, thereby setting a new benchmark in CID.

Incorporating deep learning into CID processes enhances pattern analysis and image recognition \cite{b39:deng2009imagenet}, improving accuracy and speed in large, diverse datasets. Recognizing various aircraft, especially \textit{Novel} classes, extends beyond standard image classification tasks. The primary challenge lies in the continuous introduction of new, unknown aircraft into fleets. Limited training data, diverse designs, dynamic appearance changes, and intra-class variability increase remote sensing imagery complexity. These challenges necessitate advance deep learning (DL) models to adapt to evolving aircraft technology, sparse and noisy data, and the absence of comprehensive contextual information.

\begin{figure*}[ht]
  \includegraphics[width=1.0\linewidth]{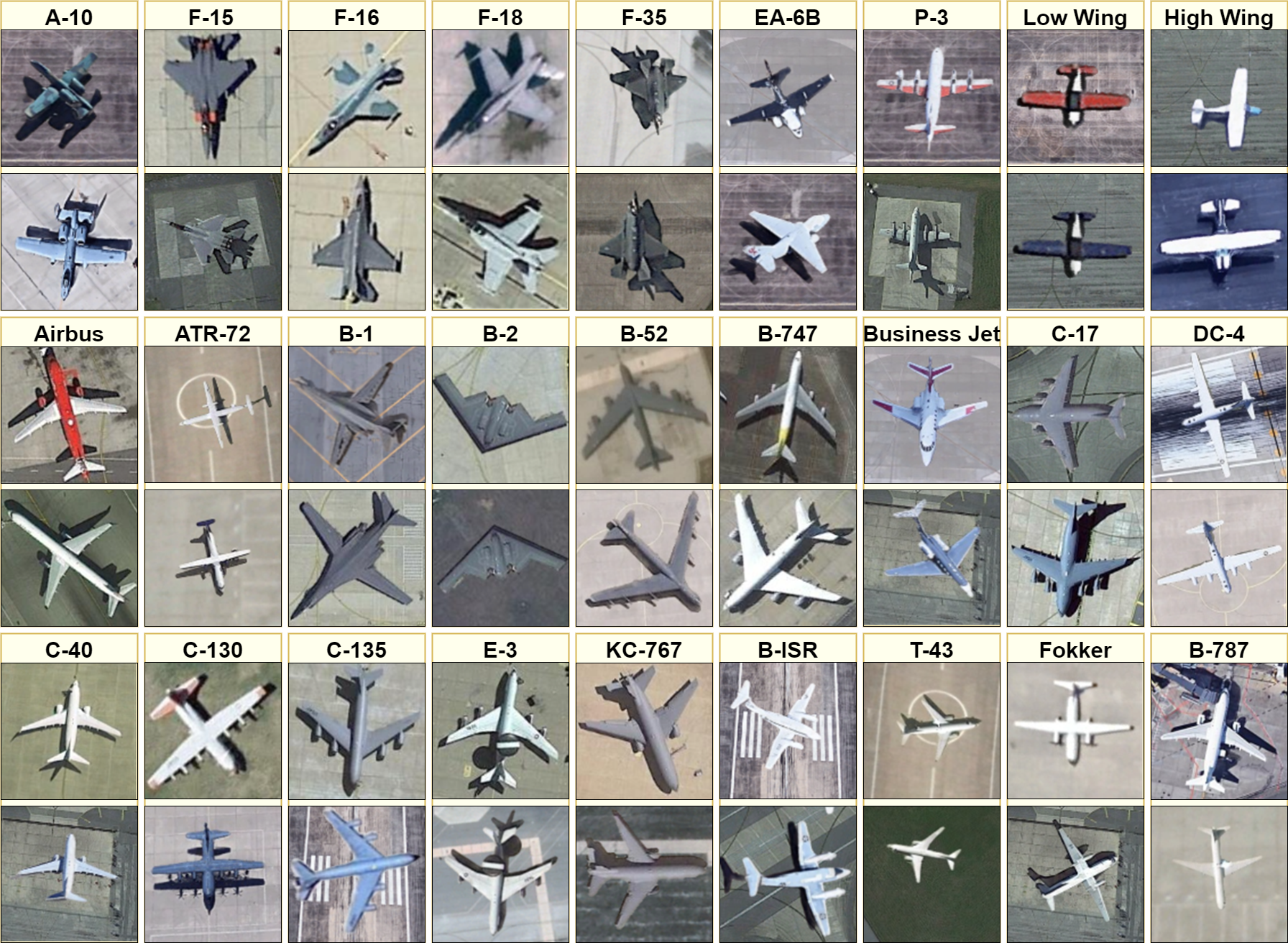}
  \caption{ MTARSI-INNAR~\cite{b47:saeed_2023_10421449} Embedder Training Set Showcasing Remote Sensing Military \& Civil Aircraft highlighting Remote Sensing challenges as  illumination conditions,orientation, background, noise and capturing angles.}
  \label{fig:datasetTrain}
\end{figure*}

To this end, we are proposing INNAR, a transformative approach that redefines the solution to the CID problem. This approach integrates a generalized aircraft embedder and similarity learning within an end-to-end deep learning framework, enabling efficient aircraft recognition with minimal training data. Current state-of-the-art multi-classification algorithms~\cite{b28:saeed2023Remote, b6:liu2021swin, b26:zhao2021aircraft},\cite{b10:guo2021research} can predict classes based on learned patterns from training data, but fall short as a comprehensive solution to the CID problem. Section~\ref{section:RelatedWork} further discusses the literature available on aircraft classification. As depicted in Figure~\ref{fig:introMethodology}, traditional CID approaches are limited to predicting classes within the domain of input training embedding, whether derived from live or offline streams. In contrast, an automated CID solution necessitates a data-efficient, generalizable, and robust approach that can recognize brand-new (few-shot samples and \textit{Novel} classes) aircraft types, even when they are absent in the training data but present in the target fleet. This approach eliminates the need for re-training or fine-tuning of the image embedder, thereby advancing the state-of-the-art in CID.

\begin{equation}
  \label{eq:pnorm}
  d(\mathbf{x}, \mathbf{y}) = \left( \sum_{i=1}^{n} |x_i - y_i|^p \right)^{\frac{1}{p}}
\end{equation}

Equation~\ref{eq:pnorm} assesses vector proximity (\( \mathbf{x} \) and \( \mathbf{y} \)) in a specified vector space. Parameter \( p \) denotes the norm type, defining distance metrics like Manhattan (\( L1 \)) and Euclidean (\( L2 \)). These feature vectors undergo processing through fully connected layers, utilizing similarity metrics crucial for security applications. The focus lies in discerning aircraft features, especially in few-shot learning. Our proposed method excels in identifying novel classes and integrates state-of-the-art techniques to create an embedding space, effectively distinguishing between \textit{Known} and \textit{Novel} classes. Section~\ref{section:Methodology} explains INNAR and its empirical evaluation.

\begin{figure}[ht]
  \centering
  \includegraphics[width=0.8\linewidth]{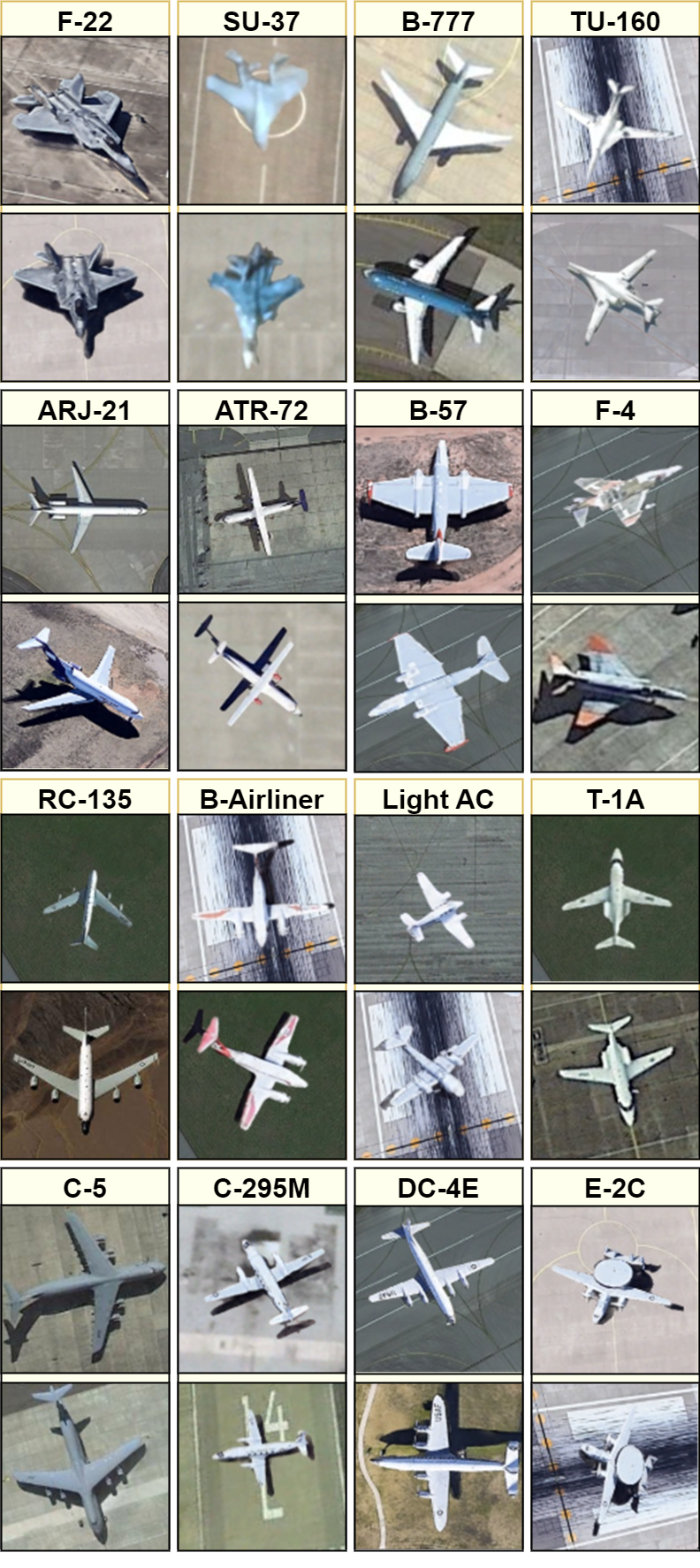}
  \caption{ MTARSI-INNAR~\cite{b47:saeed_2023_10421449}  Novel Testing Set highlighting unique classes covering role and features as Fighter, ISR, Transport, Surveillance and Multi-Mission Platforms.}
  \label{fig:datasetTest}
\end{figure}

Moreover, remote sensing imagery poses unique challenges and characteristics when compared to optical \cite{b50:HyperspectralImages},\cite{b49:TinyImages} \cite{b40:maji13fine-grained} and medical images \cite{b15:bilal2023role, b16:bilal2023aggregation, b45:ahmed2021prnet}. These distinctions arise from inherent complexities, including diverse environmental settings, limitations imposed by satellite sensors, fluctuating weather conditions, and noisy datasets as highlighted in these studies~\cite{b54:zhao2021mgml, b55:ding2102object}. In this study, we worked with the Multi-Type Aircraft Remote Sensing Images (MTARSI) dataset \cite{b43:rudd2021multi},  Figure~\ref{fig:datasetTrain} \&~\ref{fig:datasetTest} show a few samples of all aircraft types from training and testing sets showcasing diversity in capturing angles and simulating scenarios. MTARSI is the only dataset for aircraft recognition, but contains several errors and issues related to class labeling and cross-contamination. We have performed a thorough evaluation and cleaning of the MTARSI dataset \cite{b7:wu2020benchmark} by involving input from field experts (pilots) and satellite imagery analysis. This process aimed to conclusively address class labeling and cross-contamination issues, as highlighted in Figure~\ref{fig:datasetTrain}. Our refined techniques not only reduce errors but also facilitate a precise, unambiguous interpretation of the dataset, thereby establishing a new benchmark for technical precision in data processing. Section \ref{section:dataset} describes the dataset evaluation and cleaning procedure in detail.

Our INNAR approach is a novel method that effectively integrates a thresholding technique in the recognition process that effectively determines image alignment with the embedding space distribution, enabling informed decision-making during testing. Simultaneously, it incorporates advanced deep nets within the existing framework, eliminating the need for a completely new architecture, showcasing its efficiency and adaptability in addressing key challenges in aircraft recognition. Unlike image classification that makes false promise of very high accuracy (up to 99\%\cite{b29:gao2022optimizing}) on original noisy MTARSI dataset \cite{b22:wu2019multi}, the proposed INNAR methodology utilizes cutting-edge convnets and leverages similarity learning to construct robust and generalizable feature representation with image triplets.

This paper makes significant contributions to the aircraft recognition and combat identification, which are listed below:
\begin{enumerate}
  \item We introduced a novel method to achieve CID in an automated manner.
  \item We introduced a new and better version of the dataset - MTARSI-INNAR that advances automated CID development and evaluation after identifying errors in MTARSI dataset and performing a rigorous data cleaning process that involves domain experts. \footnote{"To contribute in open science and reproducibility, we will share the code repository for research purposes and the dataset is available at Zenodo.org [21]"  }
  \item We proposed INNAR, an innovative methodology that distinguishes between \textit{Known} (Friend) and \textit{Novel} (Foe) classes with high accuracy to enhance adaptability in CID. It utilizes cutting-edge convnets and leverages similarity learning to construct robust and generalizable feature representation with image triplets.
  \item We validated that image classification makes false promises of very high accuracy on original noisy MTARSI dataset \cite{b22:wu2019multi} and propose a shift to similarity and few shot learning for aircraft recognition and combat identification\label{1}.
\end{enumerate}

\section{Related Work}
\label{section:RelatedWork}
The evolution of aircraft recognition in remote sensing, with its multifaceted components of classification, detection, and segmentation, has shown a paradigm shift in recent years towards end-to-end network development. This shift underscores the increasing significance of seamless integration across diverse applications. Dire shift in from convolution\cite{b4:ucar2020aircraft},\cite{b1:he2016deep} to Transformers\cite{b5:dosovitskiy2020image} and now again convNext\cite{b52:ConvNeXt} Deep Learning networks has advanced formidably, particularly in object classification in the form of feature representation/ extraction capabilities. The contemporary approach treats object detection as a classification challenge for regions of interest, harnessing deep architectures to autonomously extract intricate image features.

In order to address aircraft shadows and robust feature extraction combination of methods are introduced like \cite{b35:wu2014aircraft} direction estimation  through a reconstruction-based similarity measure and a jigsaw matching pursuit algorithm. Evaluation with Quickbird imagery validates the method's effectiveness, while a Principal Component Analysis (PCA)-based CNN approach \cite{b34:liu2012aircraft}  showcases enhanced segmentation with a novel low downsampling ratio, though it encounters difficulties in addressing large aircraft in salient regions and managing unbalanced datasets.An impressive work in this domain Multiple Class Activation Mapping (MultiCAM) approach \cite{b27:fu2019multicam} demonstrates improved class activation maps by precisely localizing object parts and addressing background interference, yet faces challenges in covering large aircraft, ensuring interpretability with unbalanced datasets, and managing CNN mapping complexity.  However, limitations include insufficient evaluation details, concerns about generalization, and a lack of real-world examples and datasets.In the same relam, TransEffiDet \cite{b30:wang2022transeffidet},innovates aircraft detection with EfficientDet and Transformer, excelling in aerial imagery with BiFPN. Achieving enhanced military accuracy, challenges include detecting specific aircraft types due to less distinct shapes and lacking real-time detection, critical for timely responses.
An important factor in above-mentioned methods and established remote sensing datasets like UCMerced LandUse\cite{b19:yang2010bag}, NWPU-RESISC45\cite{b32:zhang2018aircraft}, PatternNet\cite{b21:zhou2018patternnet}, AID \cite{b53:AID}\cite{b20:cheng2017remote},\cite{b35:wu2014aircraft} and FAIR1M\cite{b36:sun2022fair1m} present challenges by treating aircraft as a single class, creating a skewed distribution that hinders accurate classification due to intricate complexities. This limitation makes these datasets unsuitable for \textit{Known} and \textit{Novel} class recognition techniques, revealing a critical research gap in addressing the challenges of CID and \textit{Novel} class recognition in aircraft datasets.
Aircraft recognition in eXplainable Artificial Intelligence (XAI) framework by  \cite{b31:tang2020srarnet} makes a notable contribution by combining the Hybrid Global Attribution Mapping (HGAM) algorithm, Path Aggregation Network (PANet) module for feature learning, and Class-specific Confidence Scores Mappings (CCSM) metrics to enhance deep neural network interpretability in synthetic aperture radar-based aircraft detection. Despite introducing novel XAI techniques to a domain with known challenges in comprehensibility, the method exhibits limitations such as insufficient evaluation details, concerns about generalization, and a lack of real-world examples and datasets.
\begin{figure}[ht]
\centering
  \includegraphics[width=0.8\linewidth]{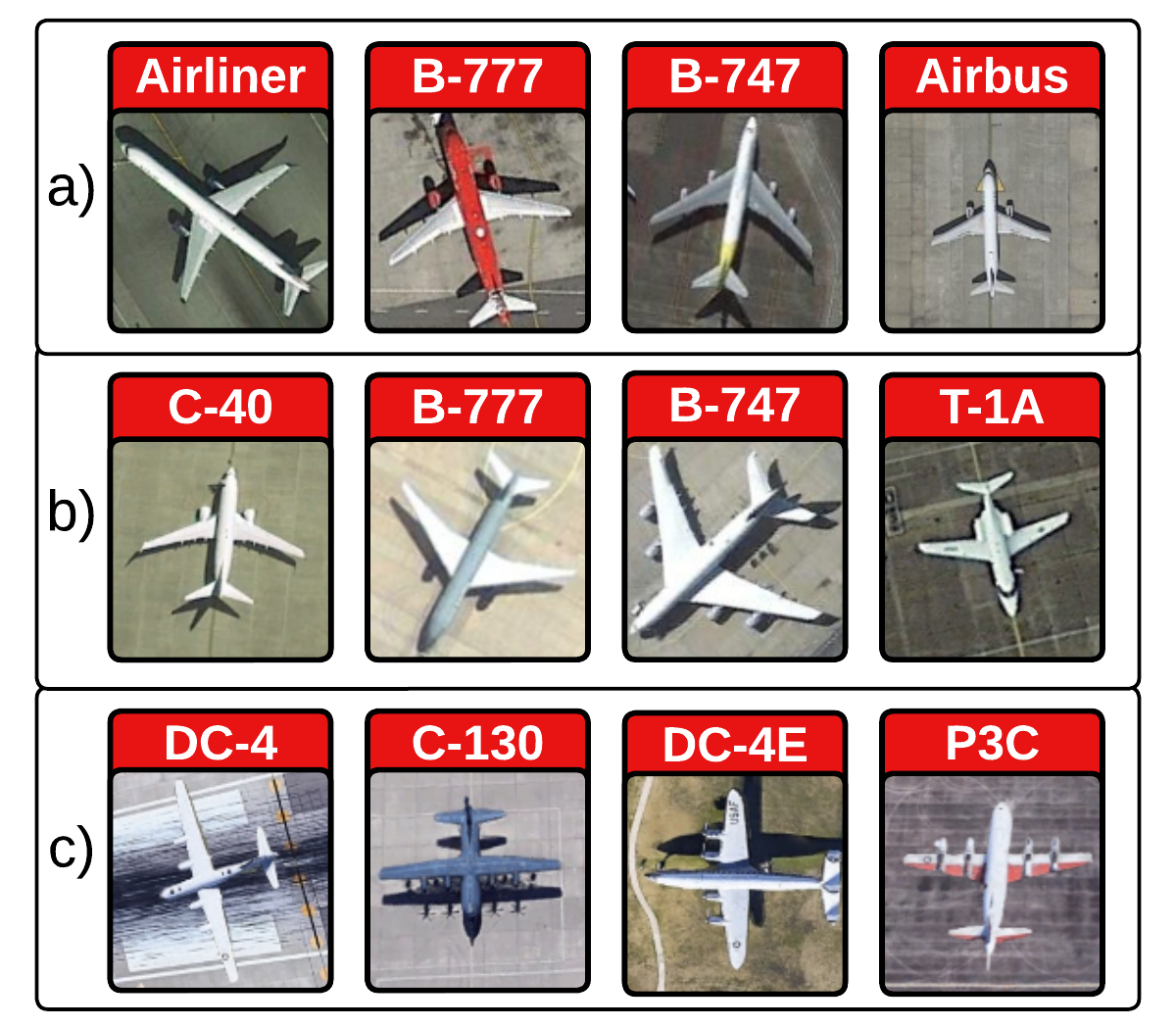}
  \caption{Miss labelling \& cross contamination issues in MTARSI datasets. a) B-777, B-747 and Airbus being displayed as Boeing. b) B-777, B-747 and T-1A being misclassified with C-40. c) C-130, DC-4E and P3C being displayed as DC-4.}
  \label{fig:datasetIssues}
\end{figure}

In 2020, \cite{b7:wu2020benchmark} addressed the lack of standardized benchmarks in aircraft type recognition by introducing the MTARSI dataset, consisting of 9,385 images of 20 aircraft types and over 12,000 samples in version -1 \& II \cite{b22:wu2019multi} \cite{b43:rudd2021multi} with diverse backgrounds and resolutions. The dataset proved crucial for fair comparisons, demonstrating a 26\% improvement in average accuracy through transfer learning using shallow networ like AlexNet and EfficiNet. However, limitations include noise and imbalance in dataset, neglecting multi-resolution impact on aerial images for generalization to diverse scenarios in remotely sensed images. The MTARSI dataset open new horizon for the researcher community on military aircraft recognition, where~\cite{b25:azam2021aircraft} introduces a novel approach integrating handcrafted and DCNN features, achieving successful identification. However, limitations include lack of impact of the multi-resolution RS images specifically negating noise and imbalance in dataset, thus more emphasis on accuracy rather than correct identification.
In the realm of advance network techniques, like Swin Transformer \cite{b29:gao2022optimizing} in combination of  state-of-the-art super resolution method for image upsampling  which improves validation accuracy and emphasizes the importance of the training procedure over model selection, suggesting a nuanced understanding of the dataset and the learning process. It also critically examines the limitations of the MTARSI dataset, such as mislabeling, class scopes, and cross-contamination, and recommends training on multiple datasets to achieve generalizable models. Another paper \cite{b28:saeed2023Remote} applies advanced deep learning methods, such as Vision Transformers, ResNet50v2, EfficientNetB0, and InceptionNetV3, to the MTARSI dataset and achieves higher classification accuracy than the benchmark. It focuses on the design and fine-tuning of the architectures to capture the features and patterns of complex military platforms.  In current era , the introduction of few shot learning by \cite{b38:li2023domain} introduces S2I-DAFSL, a novel domain adaptive few-shot learning method for aircraft recognition, utilizing an attention transferred importance-weighting network (ATIN) for enhanced transferability from satellites to inverse synthetic aperture radar (ISAR). This approach strength lies in cross-domain few-shot ISAR aircraft recognition tasks, however,it lacks in task-specific model adaptability to new samples alongwith domain adaptive robustness, assuming labeled data availability in the source domain.

Model training procedure and role of loss, is essential for effective image recognition and classification in handling imbalanced, noisy, and heterogeneous data \cite{b44:bilal2023development}. The batch hard triplet approach, which offers stability and faster convergence, outperforms conventional methods for image recognition. However, it candidly admits the approach's limited success in achieving high accuracy on $Unknown$ classes, acknowledging the persistent challenge posed by fine-grained classes with high intra-class variance. FaceNet significantly contributes \cite{b23:schroff2015facenet} by introducing a unified system for facial recognition, optimizing training to generate discriminative embeddings capable of recognizing and verifying faces under various conditions. Its innovative use of a triplet-based loss function efficiently minimizes distances between similar faces and maximizes distances between different individuals, marking a substantial leap in facial analysis and verification systems.

\section{Proposed INNAR Methodology}
\label{section:Methodology}
\begin{figure*}[ht]
%\centering
 \includegraphics[width=1.0\linewidth]{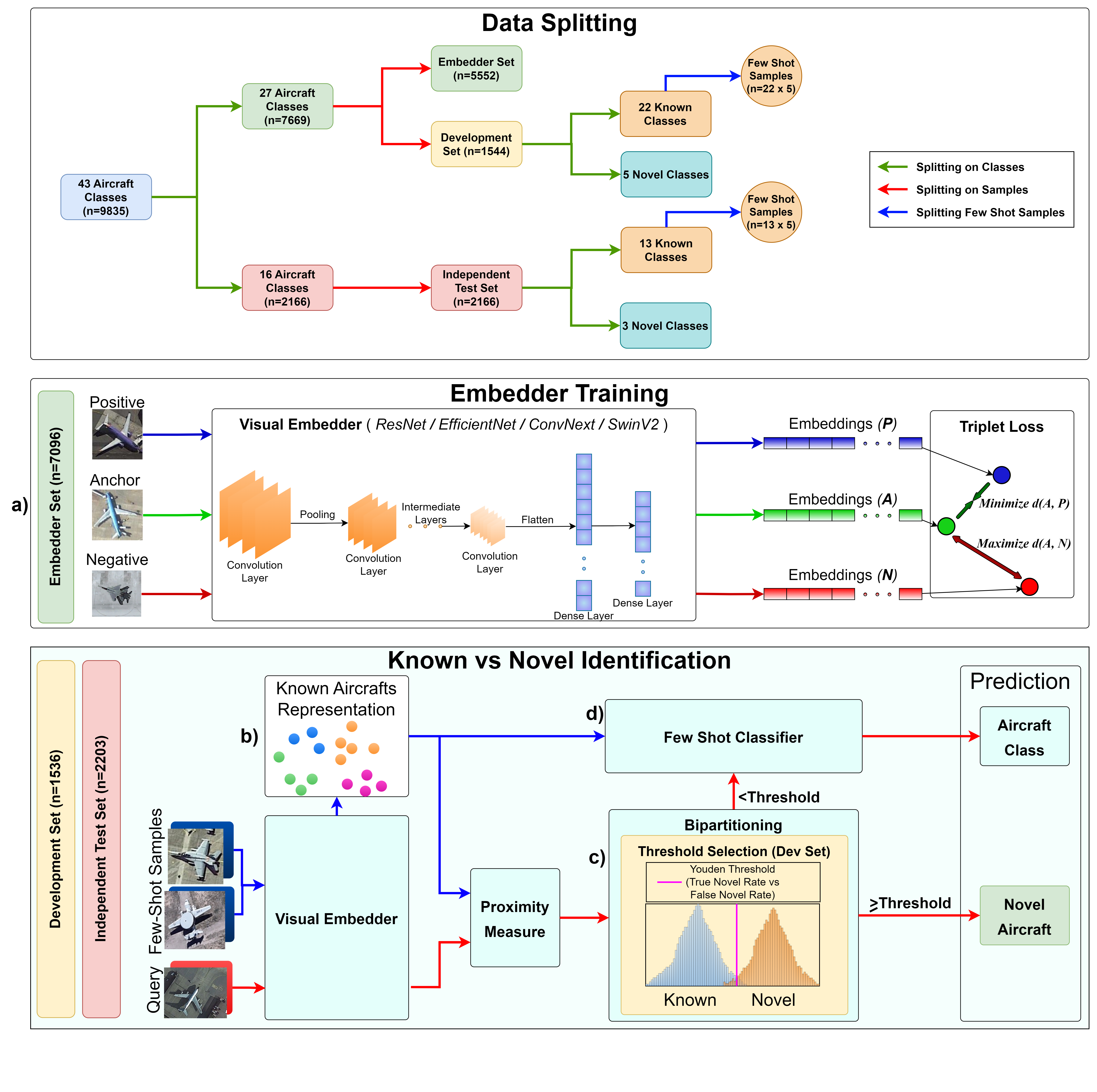}
 \caption{Proposed INNAR Methodology of Automated CID to identify Novel Class; where the input image gets transformed into feature vector which is then used to identify as a \textit{Known} or \textit{Novel} followed by classification. a) Embedder Architecture training b) Few shot learning for embedding space c) Bipartitioning for out-of distribution images d) Few-Shot Classification}
 \label{fig:proposedMethodology}
\end{figure*}

This research focuses on designing an end-to-end pipeline for aircraft recognition, particularly in identifying \textit{Known} classes vs \textit{Novel} aircraft. Firstly, we train a model to recognize aircraft via similarity learning, elaborated in Section~\ref{section:training}. This forms the basis for the next steps and is essential for the model to distinguish effectively between various aircraft categories. Secondly, we use the model trained in the first step to construct an embedding space through few-shot learning. Thirdly, we introduce a model adaptive thresholding mechanism discussed in Section~\ref{section:thresholding} to determine if an image belongs to a \textit{Known} or \textit{Novel} distribution for the recognition pipeline. If the image belongs to a \textit{Known} category, it undergoes further classification using a few-shot classifier, refining the model's discernment for more granular distinctions within established aircraft categories. To understand the process better, the details of each step and the intricate workings contributing to the model's prowess in aircraft recognition in subsequent sections and as illustrated in Figure~\ref{fig:proposedMethodology}.

\subsection{Embedding Architecture Training}
\label{section:training}
 The siamese network learns embeddings designed to capture the similarity between input triplets. During the training process, we construct triplets of samples, comprising an anchor, a positive, and a negative sample. The anchor signifies the input for which we seek to learn a meaningful embedding, with the positive being a sample similar to the anchor and the negative being a dissimilar sample. The objective actively involves minimizing the distance between the anchor and the positive while simultaneously maximizing the distance between the anchor and the negative, as highlighted in Equation~\ref{eq:tripletEq}. This approach effectively creates a margin that encourages the model to map similar instances close together and dissimilar instances far apart in the embedding space, as illustrated in the Embedder Training section within Figure~\ref{fig:proposedMethodology}.

\begin{equation}
 \label{eq:tripletEq}
 L(a,p,n) = \max\{d(a_i, p_i) - d(a_i, n_i) + margin, 0\}
\end{equation}

The loss function associated with the triplet loss typically takes the form of the hinge loss, penalizing the model when the margin constraints are violated. The network's parameters are then updated through backpropagation, optimizing the embedding space for the given task. This methodology proves especially effective in scenarios where pairwise information about similarity or dissimilarity between samples is available, making it valuable for tasks ranging from image verification to recommendation systems.

For a comprehensive evaluation, we conducted extensive trials using a diverse range of architectures, exploring variations in both structure and nature to ensure a thorough examination of the model's performance. This involved testing with several deep learning architectures like ResNet, EfficientNet, Swin Transformer, and ConvNext, each serving as embedders to generate distinct embedding vectors for our evaluation purposes.

%\subsection{Few shot learning for embedding space}
\label{section:fewShot}
After completing the initial step, we acquired an embedder with the ability to generate vector representations for any input aircraft image. To overcome data scarcity and ensure scalability for newly incorporated aircraft without retraining, we embraced few-shot learning. We considered a limited set of images for each aircraft, fed them into the embedder, and obtained corresponding embedding vectors. These vectors, along with their labels, were stored to create an embedding space addressing both data scarcity and scalability issues. This space allows us to calculate distances with test-time embedding vectors using Euclidean Distance.

\subsection{Bipartitioning for out-of distribution images}
\label{section:thresholding}
In this crucial step, we introduce a thresholding technique to discern whether an image aligns with the distribution of our generated embedding space. If the image falls outside our generated embedding space, it indicates that the provided image does not pertain to any of the aircraft classes encompassed by our space. To achieve this, we transformed the aircraft labels into a binary format, where 0 represents the classes used for our known aircraft representation, and 1 signifies classes set aside that are not part of this space. With the transformed labels, we selected an image from the development set and computed its distance within our embedding space. These distances for all development images were recorded and used to plot their distributions as highlighted in Figure~\ref{fig:histogramCombined} and compute Receiver Operating Curve (ROC) enabling the identification of the True Novel Rate (or True Positive Rate) and False Novel Rate (or False Positive Rate) respectively. By scrutinizing this, we can make an informed decision on selecting a threshold that aligns with our model and other preferences—whether prioritizing a higher TPR, a higher FPR, or a balanced point. This chosen threshold is then applied during testing to distinguish images belonging to our distribution from those in out-of-distribution classes. For this study, we opted for Youden Index for the selection of optimal threshold which tries to capture maximum potential effectiveness for a distribution. After determining whether an image falls within our embedding space. The next step involves classification for those images that do belong to the space. Employing the K-Nearest Neighbor method, we leverage the \textit{Known} embedding space and their corresponding labels for this classification. Once an embedding vector is generated for a test image, its distances are computed with all vectors in the embedding space. The labels of the closest K vectors are extracted, and a consensus is reached among these labels, assigning the most recurrent label to the image in question. To gauge the overall performance of our pipeline, we evaluate it using  weighted F1-Score.

%\subsection{Few-Shot Classification}
\label{section:classification}

This algorithm goes into detail about how the final step is carried out. It explains the specific actions and methods involved in this step, aiming to make the process more understandable. The goal is to provide a clear and accessible explanation of the final phase, ensuring that the steps taken are straightforward and easy to grasp within the overall algorithm.

\begin{algorithm}
  \caption{Automated CID Embedding-based  Algorithm}
  \begin{algorithmic}[1]
   \renewcommand{\algorithmicrequire}{\textbf{Input:}}
   \renewcommand{\algorithmicensure}{\textbf{Output:}}
   \REQUIRE Images
   \ENSURE Aircraft recognition into \textit{Known} classes or \textit{Novel} Class

   \STATE \textit{Initialization:}
   $Embedder, Threshold, K$

   $Set_A = \{x_{A1}, x_{A2}, \ldots, x_{AN}\}$ (Set of \textit{Known} Classes)
   $Set_B = \{x_{B1}, x_{B2}, \ldots, x_{BN}\}$ (Set of \textit{Novel} Classes)
   $Set_C = Set_A \cup Set_B$ (Set of All Classes)

   \STATE \textit{Few Shot Phase:}
   \FOR{each class $c$ in the $Set_A$}
     \STATE Sample $N$ images $\{x_1^c, x_2^c, \ldots, x_N^c\}$ from class $c$
     \STATE Generate embedding vectors $\{v_1^c, v_2^c, \ldots, v_N^c\}$ using Embedder
     \STATE Store vectors $(v_i^c, c)$ for $i=1$ to $N$ in the database
     \STATE Store labels $(l_i^c, c)$ for $i=1$ to $N$ in the database
   \ENDFOR

   \STATE \textit{Testing Phase:}

   \FOR{each class $c$ in the $C$}
     \FOR{each sample $s$ in the $c$}
     \STATE $v_{\text{test}}\gets Embedder(s)$
     \STATE $distances\gets $Compute $p$-norm distance between $v_{\text{test}}$ and $v_i$ for $i=1$ to $N$
      \STATE $topKsamples\gets $ Select the top $K$ samples with minimum distance
      \STATE $meanDistance\gets $ Calculate the mean of the $topKsamples$
      \IF{$meanDistance \geq Threshold$}
        \STATE \textbf{Output:} Classify as \textit{Novel}
      \ELSE
        \STATE Pass $v_{\text{test}}$ through our Few-Shot classifier
        \STATE \textbf{Output:} Final Classification Result
      \ENDIF
     \ENDFOR
   \ENDFOR
  \end{algorithmic}
\end{algorithm}
\section{Dataset Quality Management}
\label{section:dataset}

In 2020, the introduction of the MTARSI dataset \cite{b22:wu2019multi},\cite{b43:rudd2021multi}, which constitutes three versions (V1,V2 \& V3) offers a valuable resource for advancing research in remote sensing and deep learning by establishing benchmarks to compare different models. %The description of datasets along with class distribution is highlighted in the table.
However, the  datasets constitutes severe challenges like contamination, miss-classification, duplicity, labeling inaccuracies and integration inconsistencies. Certainly, Cross-contamination in image classification labels involves unintentional mixing or mislabeling of images from different classes during machine learning model training or testing, leading to reduced accuracy and performance as the model may struggle to distinguish between classes due to contaminated label information.

\begin{figure}[ht]
\centering
 \includegraphics[width=1.0\linewidth]{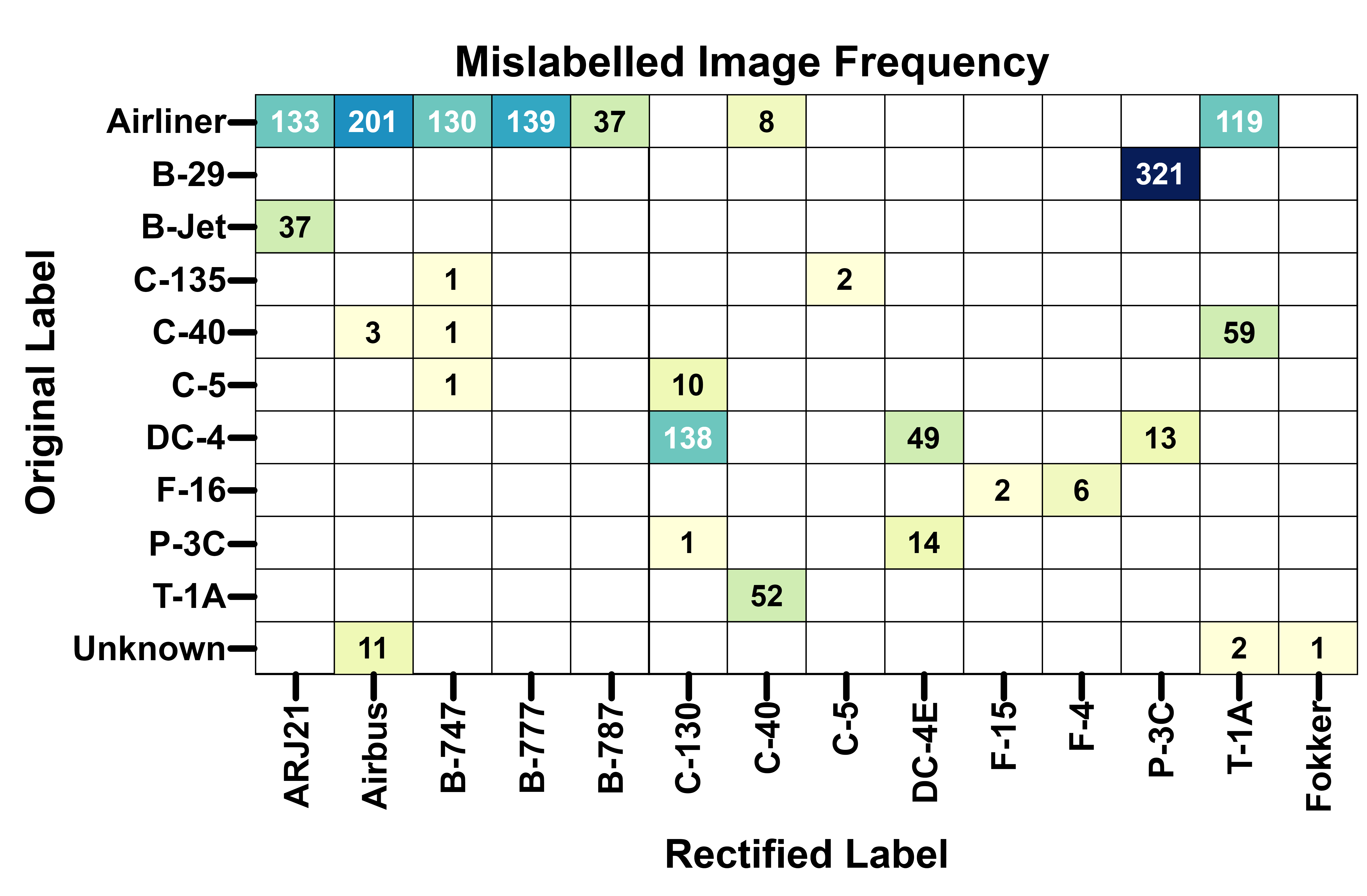}
 \caption{MTARSI-INNAR~\cite{b47:saeed_2023_10421449} samples with the number of Classes rectified for duplicity \& cross contamination in comparison to MTARSI  datasets with corresponding labels}
 \label{fig:MTARSILabelRectification}
\end{figure}

To review and clean the MTARSI\cite{b22:wu2019multi},\cite{b43:rudd2021multi} dataset, we adopt a comprehensive step by step approach, whose broader contours are highlighted as follows:-
\begin{enumerate}
\item A careful study of the dataset highlights identification of four data issues in the shape of interference by objects in the vicinity, image dimensions not compatible with the size (bigger aircraft with squeezed shapes), aircraft background noises, image merging with the same surface (e.g black) and problems such as exposure \& weather issues, underscores the challenges within the dataset, emphasizing the need for careful consideration in subsequent analyses. The image samples in Figure~\ref{fig:datasetTrain} and Figure~\ref{fig:datasetTest} highlight these issues. Moreover, the dataset \cite{b22:wu2019multi, b43:rudd2021multi} constitutes both military and civil image samples, which provides a versatile research opportunity to analyse both domains at the same time. In Figure~\ref{fig:datasetTrain} (training image samples) and Figure~\ref{fig:datasetTest} (testing image samples), codes F, B, C, and T represent fighters, bomber, transport and trainer categories, respectively.

\item Several aircraft (up to 10 different) types erroneously labelled as airliner class, which require careful re-categorization to overcome the contamination. The confusion matrix in Figure~\ref{fig:MTARSILabelRectification} reports the number of samples mislabelled because of this contamination and Figure~\ref{fig:datasetIssues} (a) show a few of those samples.
\item B-29 Superfortress and P-3\_Orion types contain highly identical image samples which we have merged together as a single class as listed in Figure~\ref{fig:MTARSILabelRectification}.
\item As illustrated in Figure~\ref{fig:datasetIssues} (b), the confusion between the C-40 and aircraft types B-777, B-747, and T-1A arises from their shared characteristics as passenger planes.
\item Figure~\ref{fig:datasetIssues} (c) emphasizes the confusion between DC-4 with C-130, DC-4E, and P3C aircraft, highlighting the inherent visual similarities that affect representation learning and classification.
\item In the training set, mislabeling occurs due to similarities between F-4, F-18, and F-22 with F-15, leading to potential convergence issues.
\end{enumerate}

We have taken following steps to clean MTARSI dataset from above-mentioned issues and duplicate samples:-
 \begin{enumerate}
 \item A dedicated team of 12 specialists, as acknowledged in Section~\ref{section:Acknowledgements}, cleaned the dataset. This team comprises professionals with diverse expertise, including pilots and experts in satellite imagery considering the role and feature types of various aircraft image samples.
 \item Duplicate samples, where the same sample is present in multiple aircraft types, are identified automatically through a Python code snippet for duplicates.
 \item We perform an iterative approach to refine the dataset, actively reviewing and updating it based on feedback from specialists and any newly acquired knowledge about the aircraft types.
 \item Finally, the above-mentioned steps ensured unified scaling to enhance quality and accuracy. Following this, the samples are normalized to 224 pixels × 224 pixels to ensure compatibility with specific models or algorithms, ensuring efficient processing and analysis.

\end{enumerate}

\section{Experimental Results And Discussion}

\begin{figure*}[ht]
 \centering
  \includegraphics[width=0.8\linewidth]{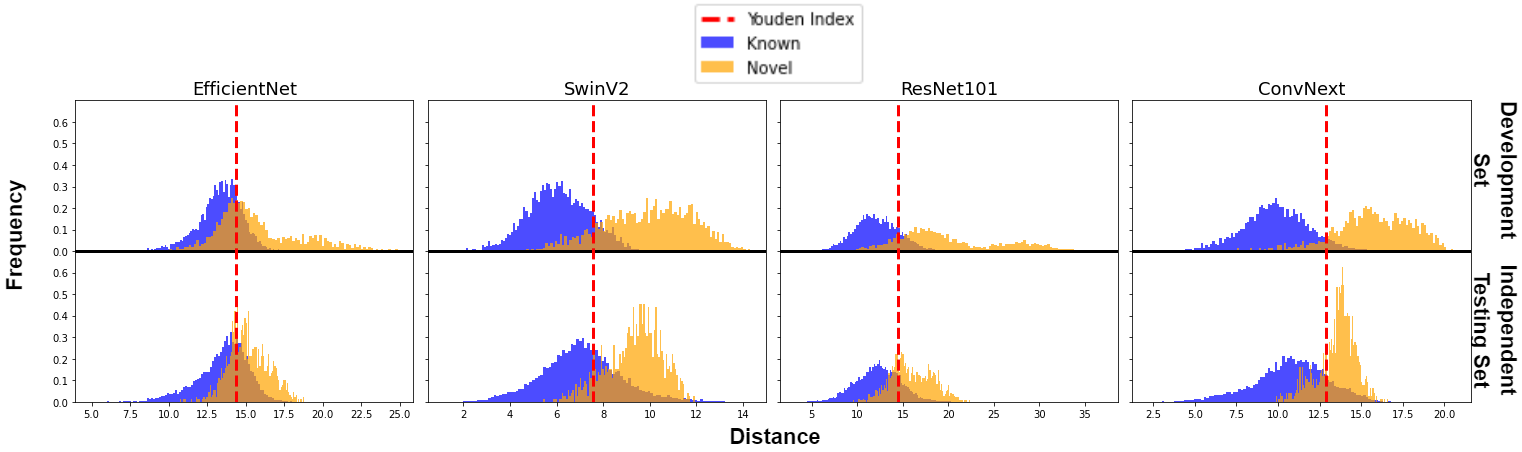}
  \caption{Histogram showing \textit{Known} vs \textit{Novel} class distributions for State of Art deep learning architectures where X -axis represent Distance \& Y-axis represent the Frequency on MTARSI-INNAR~\cite{b47:saeed_2023_10421449} Development and Independent Test Sets.}
  \label{fig:histogramCombined}
\end{figure*}

\subsection{Experimental Setup}
\begin{figure}[ht]
  \centering
  \includegraphics[width=0.8\linewidth]{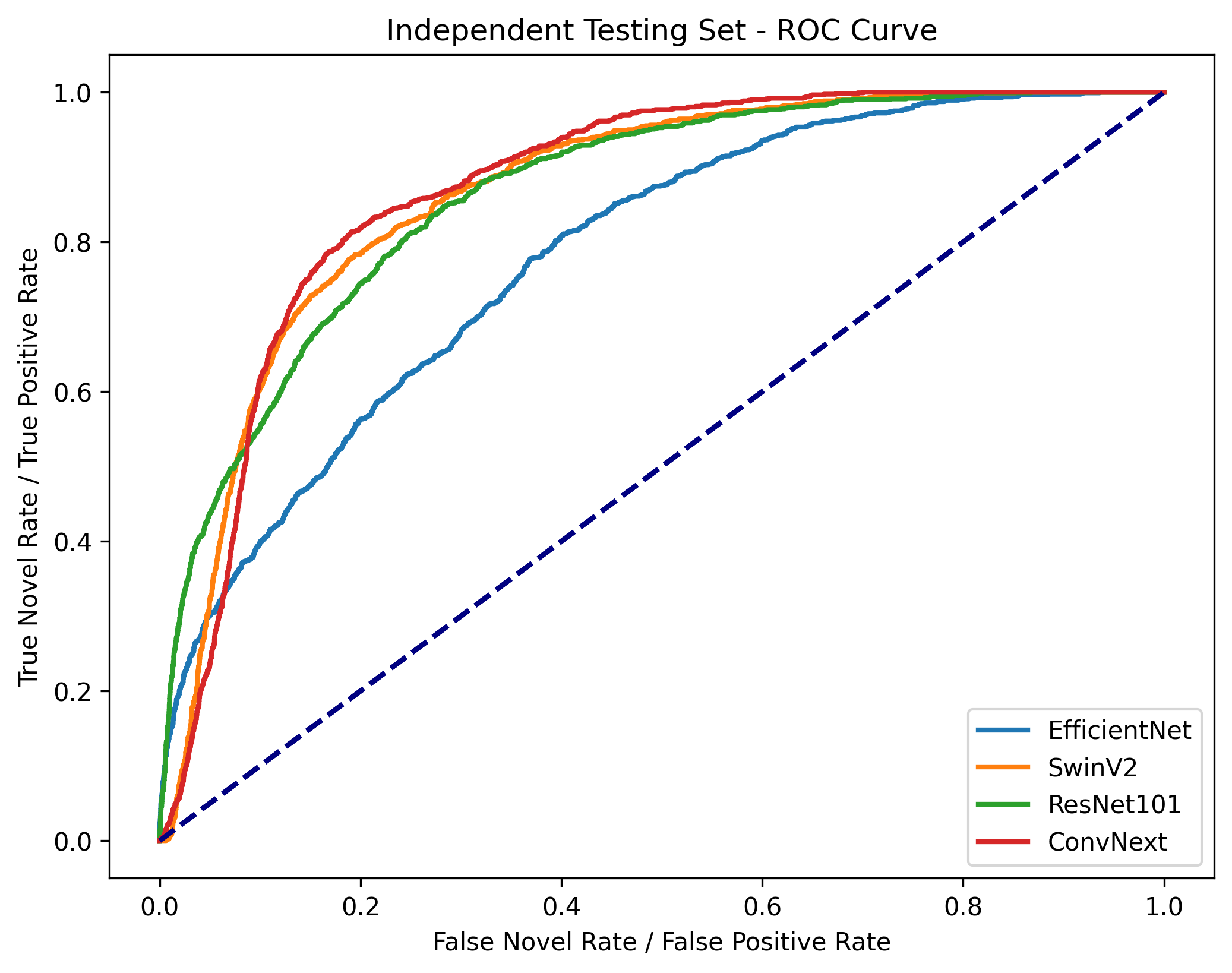}
  \caption{ROC Curve on Test Set between Known and Novel where \textit{Novel} corresponds to Positive class}
  \label{fig:rocCurve}
\end{figure}

We used Weights \& Biases, extending beyond experiment management to encompass logging, plotting, and version control. We executed the experiments on a single Nvidia Tesla P100 GPU with 16GB memory. For dataset partitioning, 80\% was allocated for training and 20\% for testing on a class-wise basis as highlighted in  where classes in bold correspond to \textit{Novel} Class. To maintain consistency, input dimensions were standardized, resizing images to 224x224 with a batch size of 32, and employing the ADAM optimizer with a learning rate of 1e-4. We utilized various architectures, including Resnet, EfficientNet, Swin, and ConvNext, as embedders. The embedder incorporated the Triplet Margin Loss, configured with a margin of 1 and a threshold derived from the median value of the ROC Curve in Figure~\ref{fig:rocCurve}. For classification, we rely on K-nearest neighbors (KNN) with a constant K value of 5 across all experiments.

\subsection{Results \& Discussion}
\label{section:discussion}
\begin{figure}[ht]
  \centering
  \includegraphics[width=0.8\linewidth]{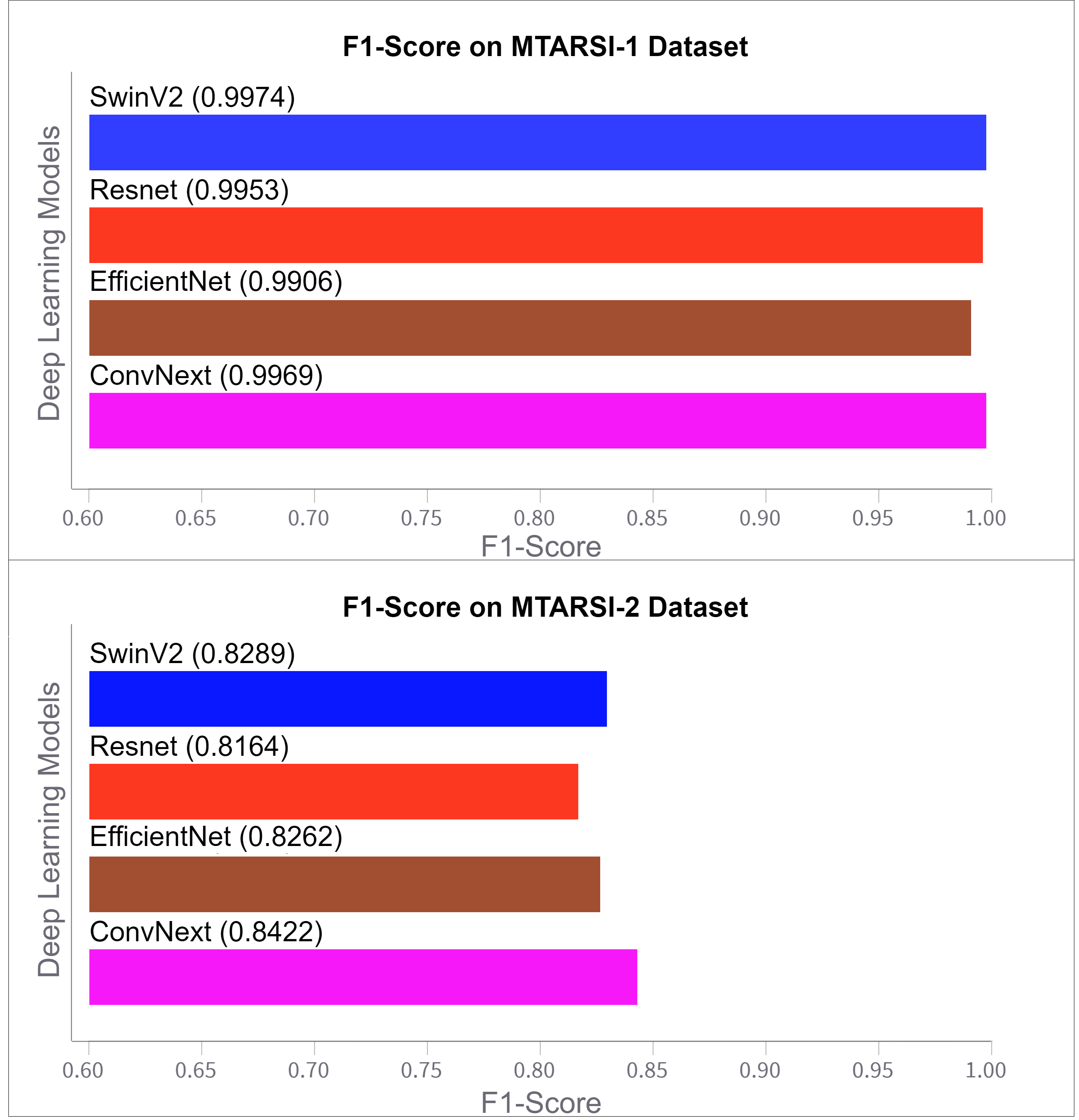}
  \caption{Comparative Analysis of MTARSI\cite{b22:wu2019multi} \& \cite{b43:rudd2021multi}. F1 Score highlights clear decline in performance of  various deep learning architectures on Noisy (left) \& less Noisy (Right) datasets}
  \label{fig:CombinedMTARSIF1}
\end{figure}

In our exploration of the \cite{b22:wu2019multi} and \cite{b43:rudd2021multi} datasets, our initial classification efforts yielded impressive results, achieving a flawless 1.0 F1-Score for \cite{b22:wu2019multi} and an 0.84 F1-Score for \cite{b43:rudd2021multi} as highlighted in Figure~\ref{fig:CombinedMTARSIF1} above. However, this success brought to light a notable challenge: the models' outstanding performance on \cite{b22:wu2019multi} revealed a significant limitation, specifically in terms of cross-contamination. This issue arose when identical images were present in both the training and testing sets of same class as well as in different class. Consequently, this prompted a critical reassessment of improved dataset to address and mitigate this unexpected phenomenon.

In the \cite{b22:wu2019multi} dataset, the challenge of overfitting is primarily driven by a complex interplay of multiple factors. These include labeling errors, inconsistent class definitions and the absence of a canonical data split. Labeling inaccuracies have proven to be a significant hurdle, causing our experimental deep learning models to memorize specific, often erroneous features within the training data. As a result, the models fixate on these unique details, failing to grasp the general characteristics crucial for accurate classification. Furthermore, the absence of a well-defined data split complicates the evaluation of a model's ability to generalize. This complicates matters by potentially leading to overfitting on the validation set, where models might perform remarkably well due to getting overly familiar to specific validation set features. The issues extend to classes like B-2, achieving perfect scores despite being indistinguishable from others. The inconsistent class definitions, as seen in the 'Boeing' class, further worsen overfitting concerns. Additionally, the F-16 class contained images unrelated to the actual F-16 aircraft, underlining the necessity of accurate data labeling and maintaining consistent class definitions for effective model training. Balancing the trade-off between correctly classifying \textit{Known} entities and accurately identifying unknown entities is a critical consideration. The specific application's requirements and priorities play a significant role in determining the threshold values.

\begin{figure*}[ht]
  \centering
  \includegraphics[width=0.8\linewidth]{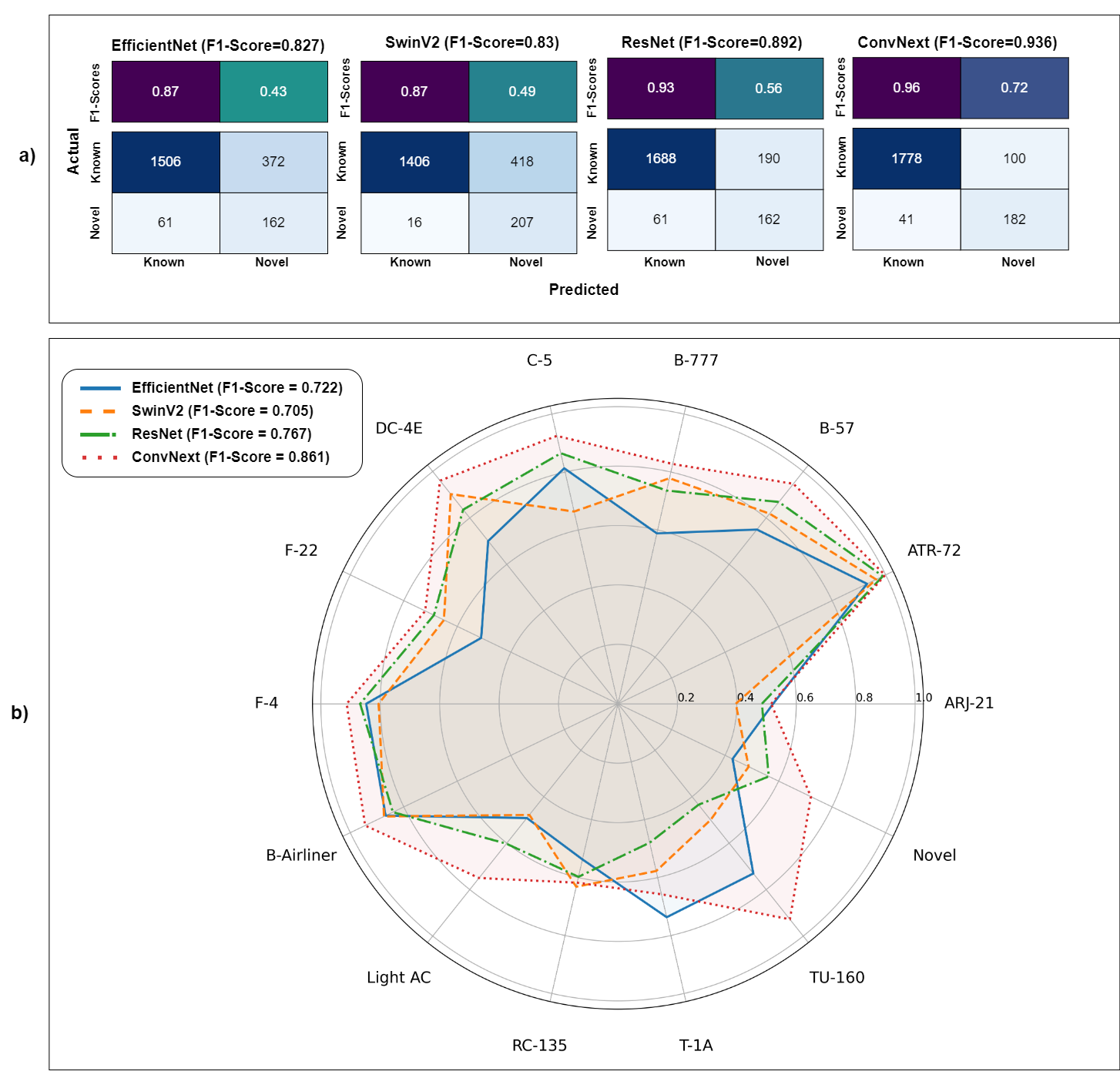}
  \caption{ Evaluation on test set of various architectures. a) Biparitioning performance visualization using confusion matrix and class-wise f1-score between Actual and Predicted labels of \textit{Known} vs \textit{Novel} on Test Set. b) Radar plot showing the performance on MTARSI-INNAR~\cite{b47:saeed_2023_10421449} Test set for 13 Known classes \& Novel class. State of art deep learning models' F-1 Score on best performing ATR-72, B-57, DC-4E aircraft but show performance decline on ARJ-21 and Light AC.}
  \label{fig:combinedEvaluation}
\end{figure*}

Addressing the challenging task of identifying \textit{Known} classes which is a crucial part of real-world problem of aircraft recognition and combat identification, our approach involved a distinctive strategy: training an embedder. This embedder, initialized with ImageNet weights, underwent further refinement through the application of the Triplet Loss Function over several epochs. In Figure~\ref{fig:histogramCombined}, we illuminate key insights into the distributions on both the development and test sets, emphasizing that each architecture embeds vectors within a uniquely ranged space. It is evident that EfficientNet and SwinV2 which are smaller architectures create the most confined embedding space (Their distributions are much compact). On the other hand, ResNet and ConvNext seems to have a wider distribution and they also have very less overlapping region between the \textit{Known} and \textit{Novel} classes which is highlighted by Blue and Orange color respectively.

\begin{figure*}[ht]
  \centering
  \includegraphics[width=0.8\linewidth]{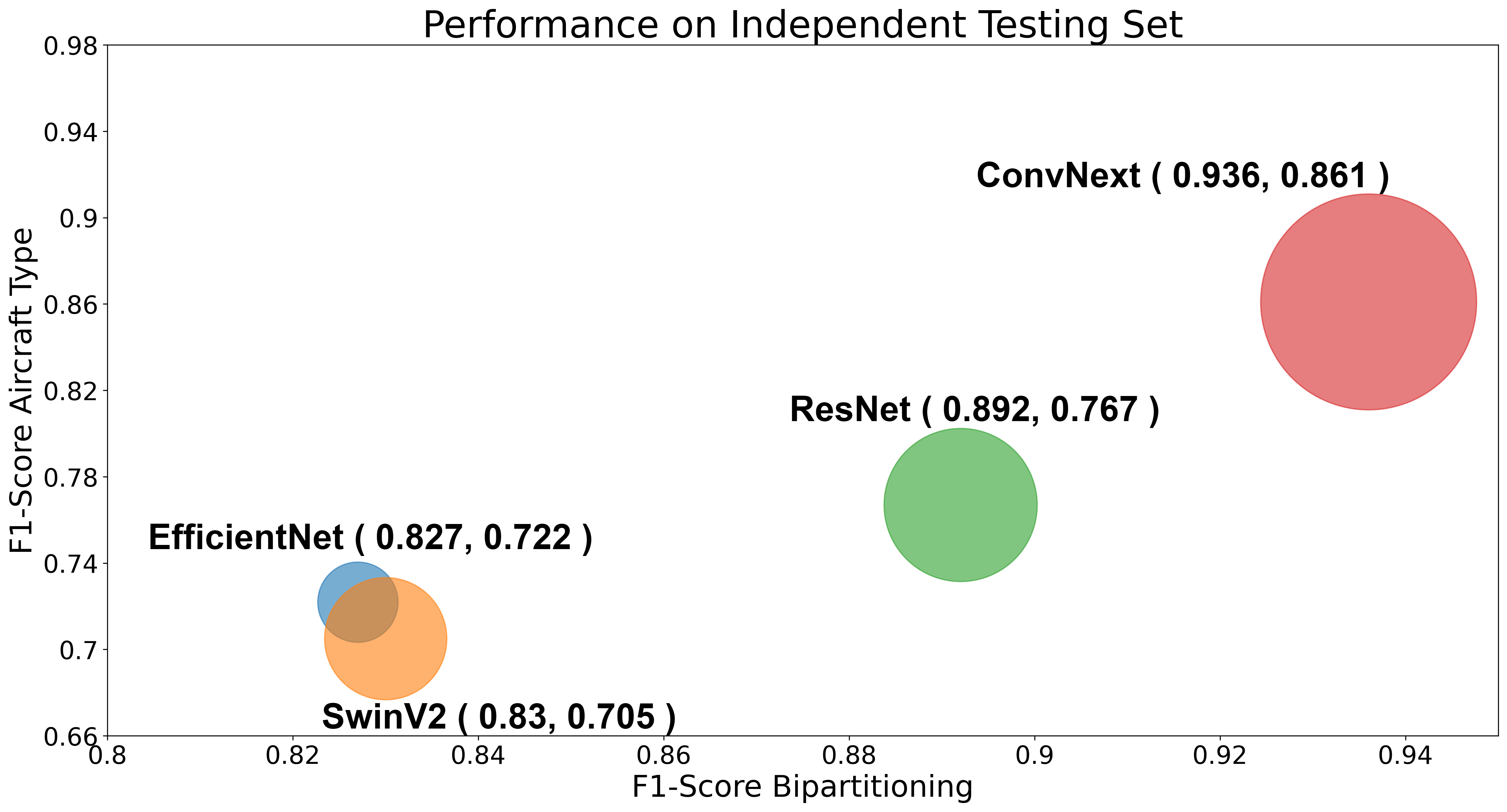}
  \caption{Performance comparison of F1-Score Classification against F1-Score Bipartitioning. The size of each point highlights the parameters of architecture}
  \label{fig:performanceComparison}
\end{figure*}

We introduced a deliberate 22-5 split to the development set as highlighted in Development Set(27 Classes), designating 22 classes as \textit{Known} and 5 as \textit{Novel}. The implementation of few-shot learning entailed selecting K=5 representative images for each of the 22 \textit{Known} classes, generating their embedding vectors, and subsequently validating them against other images from the same classes. Notably, utilizing the few shot technique on training set,  we identified an optimal threshold using Youden Index to discern the separation between \textit{Known} and \textit{Novel} classes, crucial for evaluating the performance of the embedder on the testing set. For testing set, We took this a step further by introducing 16 completely new classes, all unseen during the embedder's training. Out of these, we utilized 13 as \textit{Known} and 3 as \textit{Novel} as mentioned in Testing Set(16 Classes).

Given that threshold selection is based on the development set, certain architectures exhibit excellent performance on development data but struggle on the test set. A closer examination of SwinV2's distributions reveals a considerable separation between \textit{Known} and \textit{Novel} classes in the development set, with minimal overlap. However, this architecture undergoes a significant transformation in range and representation on the test set. The F1-Score of biparitioning on the Independent Test Set stands at 0.83.

EfficientNet, on the other hand, manifests a broader development distribution but contracts significantly on the test set however the data separation remain roughly intact. It achieves an F1-Score Bipartitioning of 0.827 on the Independent Test Set. Even though this score seems great however it is still the worst performer amongst other architectures. The largest model in our study, ConvNext, achieves the highest F1-Score Bipartitioning on the test set standing at 0.936.

ResNet showed the second best performance on bipartitioning. The distributions generated by it are much wider compared to any other architecture and it retains it's performance on both development and test set. It achieves an F1-Score bipartitioning of 0.892 on the independent test set being roughly 3\% less than ConvNext.

Each architecture exhibits unique properties on the development and test set. The confusion matrix highlights their intricacies on the Independent Test set in Figure~\ref{fig:combinedEvaluation} $a)$.

Apart from bipartitioning performance, it is also important to evaluate each architecture's performance on the classification of aircraft within the \textit{Known} Class. For this, we've computed weighted F1-Score Aircraft Types. SwinV2 performs the worst here on both the development and test set. It achieves an F1-Score of 0.847 on the development set putting which then drops down to 0.7053 on the test set. In contrast, EfficientNet is able to achieve an F1-Score of 0.7224 which is 2\% better than that of SwinV2 however the difference is not very significant. Both of these architectures fail to produce sparse embedding representations in our case. The distributions of both of these architectures are the most compact as visible in Figure~\ref{fig:histogramCombined}.

ResNet even though being a relatively outdated model than SwinV2 and EfficientNet is able to achieve an F1-Score of 0.8358 on the development set and 0.7668 on the test set. ResNet also exhibits the decline in performance which could be due to the difference of classes in embedder training set and test set. In contrast, our top-performing architecture ConvNext exhibits superior performance compared to other architectures, striking a balance between sparseness (evidenced by a higher F1-score) and generalizability across datasets. With consistent distributions and minimal overlap between \textit{Known} and \textit{Novel} classes in both training and test sets, ConvNext achieves F1-Scores of 0.8879 on the development set and 0.8612 on the testing sets revealing a modest 2\% decline. This decline is the smallest among other models in classification, indicating the model's robustness. ConvNext's balanced and consistent performance underscores its efficacy in handling both \textit{Known} and \textit{Novel} classes as well as classification.

The radar plot in Figure~\ref{fig:combinedEvaluation} $b)$ shows F1-scores obtained by each method on the test set. ResNet exhibits diverging performance pattern. It demonstrates an intense focus on specific classes, attaining remarkably high F1-Scores for them but exhibits sub-optimal performance for other classes, particularly struggling with T-1A and Light AC, where it records F1-Scores of 0.39 and 0.3, respectively. Of significance is ResNet's noteworthy competence in recognizing \textit{Novel} classes, positioning itself as the second-best performer in this regard. It is worth noting that across all architectures, there is a unanimous struggle in accurately classifying the ARJ-21 class, owing to its significant similarity to the ATR-72 — a distinction discussed in previous sections.

The challenges persist across various aircraft classes, such as Light AC, RC-135, and T-1A, where all architectures exhibit subpar performance. Nevertheless, a compelling revelation emerges in the performance of ConvNext, which maintains an F1-Score of approximately 0.72 on \textit{Novel} classes. This achievement stands out prominently within the context of our anticipated results, showcasing ConvNext's robust adaptability and effectiveness in handling previously unseen classes.

The INNAR being potentially extendable presents a well-structured and comprehensive approach to tackling image classification challenges and distinguishing between \textit{Novel} and \textit{Known} classes in the context of aircraft recognition. While the INNAR effectively addresses and adept challenges in aircraft recognition, its current pipeline has a limitation of considering image-based embedding space only and not accommodating diverse sensor (multi-modal) inputs for decision-making, like radar and infrared data that will further enhance the accuracy and impact of automated combat identification. For an effective identification of new aircraft types, INNAR needs an intermediate dataset (Development Set) with unseen aircraft types as a \textit{Novel} class; without it, determining the appropriate threshold becomes impossible. Similarly, INNAR distinguishes all unseen aircraft types as \textit{Novel} but it do not further classify each sub-type within the \textit{Novel} Category.

\section{Conclusion and Future Work}
This paper presents an innovative methodology for the recognition of military aircraft from remote sensing imagery, utilizing embedding and similarity metrics for precise few-shot learning across both \textit{Known} and \textit{Novel} aircraft classes. Our experimental evaluation, encompassing a variety of architectures of different scales and complexities, reveals a significant trend: larger models demonstrate enhanced performance in our scenario, highlighting their ability to generate sparse vector representations and capture nuanced insights effectively.

In addition to evaluating the performance of different architectures, we have addressed challenges in dataset quality, effectively mitigating issues of cross-contamination and erroneous annotations. With a focus on robustness, our approach offers a promising solution for the aviation industry and researchers striving for efficient recognition within the complex domain of military aircraft from satellite imagery. As we look to the future, we emphasize the need for diverse datasets that cover a wide range of aircraft types. These datasets not only enable more comprehensive research but also promote the development of advanced algorithms essential for accurate and nuanced aircraft type recognition. The practical implications of fine-grained classification span various domains, including aviation, defense, and other sectors where precise aircraft identification is of utmost importance.

\section* {Acknowledgments}
\label{section:Acknowledgements}
We are grateful to the pilots and remote sensing specialists (Murad Ali, Saleem Raza, Shahbaz Ahmad, and Usman Qamar) and their assistants for their expertise and dedication in refining the MTARSI\cite{b22:wu2019multi} dataset. Their careful analysis of remote sensing imagery improved the dataset’s quality \cite{b47:saeed_2023_10421449} and reliability, contributing to the progress of remote sensing.

\bibliographystyle{plain}
\bibliography{innar}

\begin{thebibliography}{10}

\bibitem{b45:ahmed2021prnet}
Salman Ahmed, Haasha bin Atif, Muhammad~Bilal Shabbir, and Hammad Naveed.
\newblock Prnet: Progressive resolution based network for radiograph based
  disease classification.
\newblock In {\em 2021 Ethics and Explainability for Responsible Data Science
  (EE-RDS)}, pages 1--5. IEEE, 2021.

\bibitem{b51:andrews2012human}
D.H. Andrews, L.C.R.P. Herz, M.B. Wolf, P.D. Harris, E.~Salas, and P.N.A.
  Stanton.
\newblock {\em Human Factors Issues in Combat Identification}.
\newblock Human Factors in Defence. Ashgate Publishing Limited, 2012.

\bibitem{b25:azam2021aircraft}
Faisal Azam, Akash Rizvi, Wazir~Zada Khan, Mohammed~Y Aalsalem, Heejung Yu, and
  Yousaf~Bin Zikria.
\newblock Aircraft classification based on pca and feature fusion techniques in
  convolutional neural network.
\newblock {\em IEEE Access}, 9:161683--161694, 2021.

\bibitem{b16:bilal2023aggregation}
Mohsin Bilal, Robert Jewsbury, Ruoyu Wang, Hammam~M AlGhamdi, Amina Asif, Mark
  Eastwood, and Nasir Rajpoot.
\newblock An aggregation of aggregation methods in computational pathology.
\newblock {\em Medical Image Analysis}, page 102885, 2023.

\bibitem{b15:bilal2023role}
Mohsin Bilal, Mohammed Nimir, David Snead, Graham~S Taylor, and Nasir Rajpoot.
\newblock Role of ai and digital pathology for colorectal immuno-oncology.
\newblock {\em British Journal of Cancer}, 128(1):3--11, 2023.

\bibitem{b44:bilal2023development}
Mohsin Bilal, Yee~Wah Tsang, Mahmoud Ali, Simon Graham, Emily Hero, Noorul
  Wahab, Katherine Dodd, Harvir Sahota, Shaobin Wu, Wenqi Lu, et~al.
\newblock Development and validation of artificial intelligence-based
  prescreening of large-bowel biopsies taken in the uk and portugal: a
  retrospective cohort study.
\newblock {\em The Lancet Digital Health}, 5(11):e786--e797, 2023.

\bibitem{b20:cheng2017remote}
Gong Cheng, Junwei Han, and Xiaoqiang Lu.
\newblock Remote sensing image scene classification: Benchmark and state of the
  art.
\newblock {\em Proceedings of the IEEE}, 105(10):1865--1883, 2017.

\bibitem{b39:deng2009imagenet}
Jia Deng, Wei Dong, Richard Socher, Li-Jia Li, Kai Li, and Li~Fei-Fei.
\newblock Imagenet: A large-scale hierarchical image database.
\newblock In {\em 2009 IEEE conference on computer vision and pattern
  recognition}, pages 248--255. Ieee, 2009.

\bibitem{b55:ding2102object}
Jian Ding, Nan Xue, Gui-Song Xia, Xiang Bai, Wen Yang, Michael~Ying Yang, Serge
  Belongie, Jiebo Luo, Mihai Datcu, Marcello Pelillo, and Liangpei Zhang.
\newblock Object detection in aerial images: A large-scale benchmark and
  challenges.
\newblock {\em IEEE Transactions on Pattern Analysis and Machine Intelligence},
  44(11):7778--7796, 2022.

\bibitem{b5:dosovitskiy2020image}
Alexey Dosovitskiy, Lucas Beyer, Alexander Kolesnikov, Dirk Weissenborn,
  Xiaohua Zhai, Thomas Unterthiner, Mostafa Dehghani, Matthias Minderer, Georg
  Heigold, Sylvain Gelly, et~al.
\newblock An image is worth 16x16 words: Transformers for image recognition at
  scale.
\newblock {\em arXiv preprint arXiv:2010.11929}, 2020.

\bibitem{b27:fu2019multicam}
Kun Fu, Wei Dai, Yue Zhang, Zhirui Wang, Menglong Yan, and Xian Sun.
\newblock Multicam: Multiple class activation mapping for aircraft recognition
  in remote sensing images.
\newblock {\em Remote sensing}, 11(5):544, 2019.

\bibitem{b29:gao2022optimizing}
Kyle Gao, Hongjie He, Dening Lu, Linlin Xu, Lingfei Ma, and Jonathan Li.
\newblock Optimizing and evaluating swin transformer for aircraft
  classification: Analysis and generalizability of the mtarsi dataset.
\newblock {\em IEEE Access}, 10:134427--134439, 2022.

\bibitem{b50:HyperspectralImages}
Zhang.~J. Gu.~Y., Zhang.~Y.
\newblock Integration of spatial–spectral information for resolution
  enhancement in hyperspectral images.
\newblock {\em IEEE Trans. Geosci. Remote Sensing}, 2008.

\bibitem{b10:guo2021research}
Yonghui Guo, Yuntao Li, and Yu~Zhang.
\newblock Research on mimo-isar high resolution imaging technology.
\newblock In {\em 2021 IEEE 4th Advanced Information Management, Communicates,
  Electronic and Automation Control Conference (IMCEC)}, volume~4, pages
  169--174. IEEE, 2021.

\bibitem{b1:he2016deep}
Kaiming He, Xiangyu Zhang, Shaoqing Ren, and Jian Sun.
\newblock Deep residual learning for image recognition.
\newblock In {\em Proceedings of the IEEE conference on computer vision and
  pattern recognition}, pages 770--778, 2016.

\bibitem{b49:TinyImages}
G.~Krizhevsky, A.;~Hinton.
\newblock Learning multiple layers of features from tiny images.
\newblock {\em University of Toronto: Toronto, ON, Canada}, 2009.

\bibitem{b38:li2023domain}
Binquan Li, Yuan Yao, and Qiao Wang.
\newblock Domain adaptive few-shot learning for isar aircraft recognition with
  transferred attention and weighting importance.
\newblock {\em Electronics}, 12(13):2909, 2023.

\bibitem{b34:liu2012aircraft}
Ge~Liu, Xian Sun, Kun Fu, and Hongqi Wang.
\newblock Aircraft recognition in high-resolution satellite images using
  coarse-to-fine shape prior.
\newblock {\em IEEE Geoscience and Remote Sensing Letters}, 10(3):573--577,
  2012.

\bibitem{b6:liu2021swin}
Ze~Liu, Yutong Lin, Yue Cao, Han Hu, Yixuan Wei, Zheng Zhang, Stephen Lin, and
  Baining Guo.
\newblock Swin transformer: Hierarchical vision transformer using shifted
  windows.
\newblock In {\em Proceedings of the IEEE/CVF international conference on
  computer vision}, pages 10012--10022, 2021.

\bibitem{b40:maji13fine-grained}
Subhransu Maji, Esa Rahtu, Juho Kannala, Matthew Blaschko, and Andrea Vedaldi.
\newblock Fine-grained visual classification of aircraft.
\newblock {\em arXiv preprint arXiv:1306.5151}, 2013.

\bibitem{b43:rudd2021multi}
R~Rudd-Orthner and L~Mihaylova.
\newblock Multi-type aircraft of remote sensing images: Mtarsi 2.
\newblock {\em Zenodo}, 2021.

\bibitem{b28:saeed2023Remote}
Ahmad Saeed, Haasha~Bin Atif, Usman Habib, and Mohsin Bilal.
\newblock Remote sensing aircraft classification harnessing deep learning
  advancements.
\newblock In {\em 2023 18th International Conference on Emerging Technologies
  (ICET)}, pages 50--55, 2023.

\bibitem{b47:saeed_2023_10421449}
Ahmad Saeed, Haasha Bin~Atif, Mohsin Bilal, and Usman Habib.
\newblock Mtarsi-innar.
\newblock {\em Zenodo.org}, January 2024.

\bibitem{b23:schroff2015facenet}
Florian Schroff, Dmitry Kalenichenko, and James Philbin.
\newblock Facenet: A unified embedding for face recognition and clustering.
\newblock In {\em Proceedings of the IEEE conference on computer vision and
  pattern recognition}, pages 815--823, 2015.

\bibitem{b52:ConvNeXt}
Tao.Lei Song.F, Peng.Z.
\newblock Raih-det: An end-to-end rotated aircraft and aircraft head detector
  based on convnext and cyclical focal loss in optical remote sensing images.
\newblock {\em IEEE Trans. Geosci. Remote Sensing}, 2023.

\bibitem{b36:sun2022fair1m}
Xian Sun, Peijin Wang, Zhiyuan Yan, Feng Xu, Ruiping Wang, Wenhui Diao, Jin
  Chen, Jihao Li, Yingchao Feng, Tao Xu, et~al.
\newblock Fair1m: A benchmark dataset for fine-grained object recognition in
  high-resolution remote sensing imagery.
\newblock {\em ISPRS Journal of Photogrammetry and Remote Sensing},
  184:116--130, 2022.

\bibitem{b46:tan2023research}
Bin Tan, Qiuni Li, Tingliang Zhang, and Hui Zhao.
\newblock The research of air combat intention identification method based on
  bilstm+ attention.
\newblock {\em Electronics}, 12(12):2633, 2023.

\bibitem{b31:tang2020srarnet}
Wei Tang, Chenwei Deng, Yuqi Han, Yun Huang, and Baojun Zhao.
\newblock Srarnet: A unified framework for joint superresolution and aircraft
  recognition.
\newblock {\em IEEE Journal of Selected Topics in Applied Earth Observations
  and Remote Sensing}, 14:327--336, 2020.

\bibitem{b4:ucar2020aircraft}
Ferhat Ucar, Besir Dandil, and Fikret Ata.
\newblock Aircraft detection system based on regions with convolutional neural
  networks.
\newblock {\em International Journal of Intelligent Systems and Applications in
  Engineering}, 8(3):147--153, 2020.

\bibitem{b30:wang2022transeffidet}
Yanfeng Wang, Tao Wang, Xin Zhou, Weiwei Cai, Runmin Liu, Meigen Huang, Tian
  Jing, Mu~Lin, Hua He, Weiping Wang, et~al.
\newblock Transeffidet: aircraft detection and classification in aerial images
  based on efficientdet and transformer.
\newblock {\em Computational Intelligence and Neuroscience}, 2022, 2022.

\bibitem{b35:wu2014aircraft}
Qichang Wu, Hao Sun, Xian Sun, Daobing Zhang, Kun Fu, and Hongqi Wang.
\newblock Aircraft recognition in high-resolution optical satellite remote
  sensing images.
\newblock {\em IEEE Geoscience and Remote Sensing Letters}, 12(1):112--116,
  2014.

\bibitem{b22:wu2019multi}
Z~Wu.
\newblock Multi-type aircraft of remote sensing images: Mtarsi.
\newblock {\em Zenodo. org}, 10, 2019.

\bibitem{b7:wu2020benchmark}
Zhi-Ze Wu, Shou-Hong Wan, Xiao-Feng Wang, Ming Tan, Le~Zou, Xin-Lu Li, and Yan
  Chen.
\newblock A benchmark data set for aircraft type recognition from remote
  sensing images.
\newblock {\em Applied Soft Computing}, 89:106132, 2020.

\bibitem{b53:AID}
Gui-Song Xia, Jingwen Hu, Fan Hu, Baoguang Shi, Xiang Bai, Yanfei Zhong,
  Liangpei Zhang, and Xiaoqiang Lu.
\newblock Aid: A benchmark data set for performance evaluation of aerial scene
  classification.
\newblock {\em IEEE Transactions on Geoscience and Remote Sensing},
  55(7):3965--3981, 2017.

\bibitem{b19:yang2010bag}
Yi~Yang and Shawn Newsam.
\newblock Bag-of-visual-words and spatial extensions for land-use
  classification.
\newblock In {\em Proceedings of the 18th SIGSPATIAL international conference
  on advances in geographic information systems}, pages 270--279, 2010.

\bibitem{b32:zhang2018aircraft}
Yuhang Zhang, Hao Sun, Jiawei Zuo, Hongqi Wang, Guangluan Xu, and Xian Sun.
\newblock Aircraft type recognition in remote sensing images based on feature
  learning with conditional generative adversarial networks.
\newblock {\em Remote Sensing}, 10(7):1123, 2018.

\bibitem{b26:zhao2021aircraft}
Baojun Zhao, Wei Tang, Yu~Pan, Yuqi Han, and Wenzheng Wang.
\newblock Aircraft type recognition in remote sensing images: Bilinear
  discriminative extreme learning machine framework.
\newblock {\em Electronics}, 10(17):2046, 2021.

\bibitem{b54:zhao2021mgml}
Qi~Zhao, Shuchang Lyu, Yuewen Li, Yujing Ma, and Lijiang Chen.
\newblock Mgml: Multigranularity multilevel feature ensemble network for remote
  sensing scene classification.
\newblock {\em IEEE Transactions on Neural Networks and Learning Systems},
  2021.

\bibitem{b21:zhou2018patternnet}
Weixun Zhou, Shawn Newsam, Congmin Li, and Zhenfeng Shao.
\newblock Patternnet: A benchmark dataset for performance evaluation of remote
  sensing image retrieval.
\newblock {\em ISPRS journal of photogrammetry and remote sensing},
  145:197--209, 2018.

\end{thebibliography}

\end{document}